\documentclass[letterpaper, 10 pt, conference]{ieeeconf}  

\IEEEoverridecommandlockouts                              

\usepackage{amsmath} 
\usepackage{amsfonts}
\usepackage{graphicx}
\usepackage{subfigure}
\usepackage{multirow}
\usepackage{cite}
\usepackage{caption}
\usepackage{float} 
\usepackage{subcaption}
\usepackage{authblk}
\usepackage{stfloats}
\usepackage{pifont}

\title{\LARGE \bf
Mr. Virgil: Learning Multi-robot Visual-range Relative Localization
}

\author{Si Wang$^{1}$, Zhehan Li$^{2}$, Jiadong Lu$^{2}$, Rong Xiong$^{1}$, Yanjun Cao$^{2}$, Yue Wang$^{1*}$
\thanks{This work was supported in part by the National Nature Science Foundation of China under Grant 62373322 and Research and Development Project of Zhejiang Province under Grant No. 2025C01205(SD2).}%
\thanks{$^{1}$Institute of Cyber-Systems and Control, Zhejiang University, China.}%
\thanks{$^{2}$Huzhou Institute of Zhejiang University, Huzhou, China.}%
\thanks{$^{*}$\textit{Corresponding author: Yue Wang.} (E-mail: ywang24@zju.edu.cn)}
}

\begin{document}

\maketitle
\thispagestyle{empty}
\pagestyle{empty}

\begin{abstract}
Ultra-wideband (UWB)-vision fusion localization has achieved extensive applications in the domain of multi-agent relative localization. The challenging matching problem between robots and visual detection renders existing methods highly dependent on identity-encoded hardware or delicate tuning algorithms. Overconfident yet erroneous matches may bring about irreversible damage to the localization system. To address this issue, we introduce Mr. Virgil, an end-to-end learning multi-robot visual-range relative localization framework, consisting of a graph neural network for data association between UWB rangings and visual detections, and a differentiable pose graph optimization (PGO) back-end. The graph-based front-end supplies robust matching results, accurate initial position predictions, and credible uncertainty estimates, which are subsequently integrated into the PGO back-end to elevate the accuracy of the final pose estimation. Additionally, a decentralized system is implemented for real-world applications.
Experiments spanning varying robot numbers, simulation and real-world, occlusion and non-occlusion conditions showcase the stability and exactitude under various scenes compared to conventional methods.
Our code is available at: https://github.com/HiOnes/Mr-Virgil.

\end{abstract}

\section{Introduction}
Relative localization is fundamental to multi-robot applications involving drone swarms, rescue missions and exploration tasks. 
One straightforward method to obtain relative estimation is the transformation of robots' global states measured by external devices such as global positioning system (GPS)\cite{jaimes2008approach,qi2020cooperative}, real-time kinematic positioning system (RTK)\cite{moon2016outdoor}, motion capture system (MCS)\cite{preiss2017crazyswarm} and multi-fixed-anchor UWB system\cite{ledergerber2015robot}. Due to the strong reliance on external infrastructure and the workspace, such solutions cannot be directly applicable to unfamiliar scenes.

To enhance scalability and accuracy, local odometry and mutual observations are integrated into the multi-agent systems, among which visual and UWB system\cite{xun2023crepes,xu2020decentralized,xu2022omni,xu2024d} serves as typical representatives. Visual images offer neighbor observations and can be leveraged for ego-motion estimation. UWB provides omnidirectional and occlusion-resistant ranging measurements. However, the noisy UWB signals\cite{li2023continuous, zhao2022finding,zhao2022util} and the drifting nature of visual odometry (VO) impose high demands on the fusion manner of multimodal information. Moreover, the visual detection targets are anonymous, rendering the correspondences between visual detection and UWB range a challenging data association problem. In order to address this, CREPES\cite{xun2023crepes} adopts active infrared (IR) LEDs and an IR fish-eye camera to achieve identity extraction, while IR communication makes time synchronization quite cumbersome. Apart from the ID-encoded hardware approaches\cite{xun2023crepes,olson2011apriltag,yan2019active}, many researches adhere to detection-based paradigm\cite{xu2020decentralized,xu2022omni,xu2024d}, which generates raw matches from visual detection bounding boxes, followed by hand-crafted rule-based post-processing. In general, these methods assume the resultant matches are correct and hard, making the downstream optimization fragile and irrecoverable when matching errors occur. Given the complexity of drone swarm formations, the ambiguity of robots and detection targets remains a challenging issue in multi-agent systems. 


To overcome the aforementioned challenges of matching aliasing, accurate uncertainty estimates should be provided alongside data association, enabling soft constraints for subsequent optimization. Additionally, matching uncertainty should be informed by the collective structure of the multi-robot formation, rather than being limited to pairwise similarity. Beyond that, in multi-robot systems, the number of robots and visual mutual observations frequently varies, raising demand for the flexibility of the matching architecture.

In this paper, we propose Mr. Virgil, an end-to-end multi-robot visual-range relative localization system. 
For the front-end network, to accommodate any number of drones and visual observations, we harness the graph neural network (GNN), which has exhibited remarkable performance in the field of image feature matching\cite{sarlin2020superglue,roessle2023end2end}.
The GNN and the differentiable Sinkhorn algorithm\cite{cuturi2013sinkhorn,peyre2019computational,sinkhorn1967concerning} are utilized to solve the data association problem, yielding matching and uncertainty estimations with a global perspective.
To supervise covariance, a Maximum Likelihood (ML) loss is applied to the output of the front-end network. The covariance is also incorporated as weights in the PGO back-end, enabling end-to-end learning using gradients from localization errors. 
The differentiable PGO back-end is employed for joint optimization and its gradient back-propagates to facilitate learning of front-end network.
To achieve real-time performance in real-world applications, we also implement a decentralized system based on Robot Operating System (ROS), LibTorch and Ceres Solver. 
Overall, our major contributions are as follows:

\begin{itemize}
  \item We propose an end-to-end decentralized multi-robot relative localization system, comprising a learnable front-end for data association and a differentiable PGO back-end.
  \item We present a GNN-based match network for multi-robot data association, realizing precise matching and reasonable uncertainty estimation, with the ability to handle an arbitrary number of drones and detections.
  \item The robustness and accuracy of our method have been verified in both simulated and real-world, occluded and non-occluded scenarios. Our network demonstrates impressive generalization ability in scenarios with limited training data, variable robot counts, and sim-to-real experiments.
\end{itemize}


\begin{figure*}[tp]
\setlength{\belowcaptionskip}{-0.6cm}
    \centering
    \includegraphics[scale=0.42]{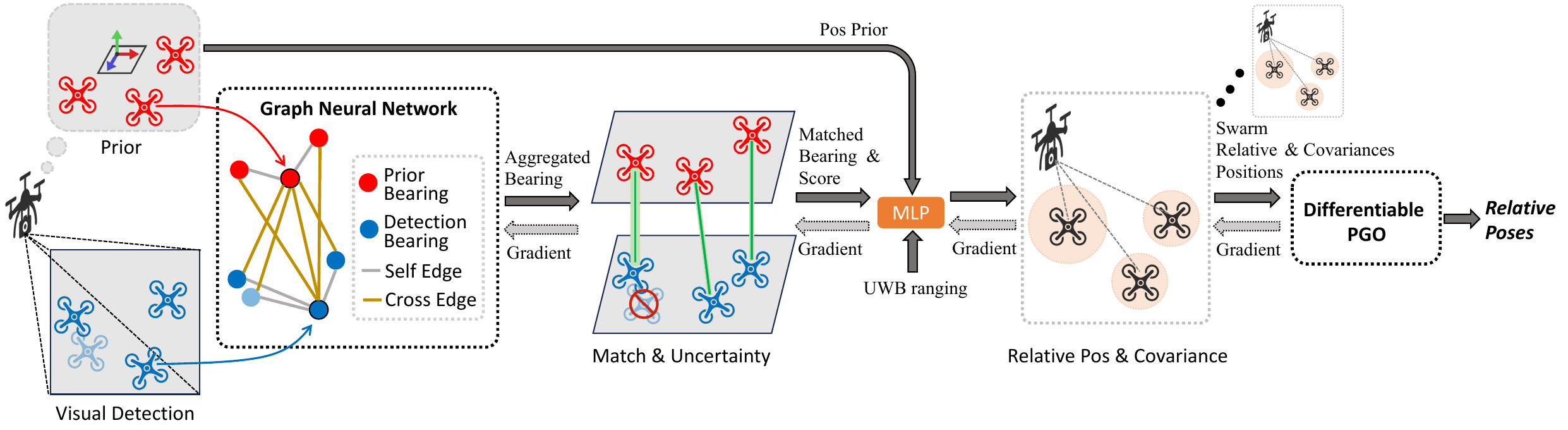}
    \caption{The pipeline of our end-to-end multi-robot localization network. The GNN-based match net associates prior bearings and detection bearings, predicting relative positions with covariances (uncertainties of matches and positions are represented by light green lines and light pink circles). The differentiable PGO improves performance and propagates gradients back for joint error correction. }
    \label{fig:pipeline}
\end{figure*}

\section{Related Works}
In this section, we discuss relative localization schemes that function without external aids like GPS or RTK, which offer superior adaptability to unfamiliar environments. Based on the sensor types, we classify the previous works into UWB-based methods and vision-based methods.

\subsection{UWB-based Methods} 
A few works treat UWB as a stand-alone localization solution, owing to its cost-effectiveness and ability to offer omnidirectional ranging. For UWB systems that do not rely on external fixed anchors, each robot estimates the positions of the neighbors within its local frame, eliminating the need for a unified global coordinate system. Zhou\cite{zhou2008robot} estimated the 3-DoF relative pose transformation between planar robots by leveraging inter-robot distance measurements and displacement estimation, requiring theoretical minimum UWB measurements. Fishberg\cite{fishberg2022multi} investigated the impact of multi-UWB tag antenna occlusion and interference on ranging errors. By applying occlusion-related weighted factors to nonlinear least squares, they achieved comparable accuracy to systems relying on continuous odometry exchange. Since only ranging information is shared, these methods incur a low communication overhead, while their stabilities are easily plagued by the noise-prone nature of UWB.

\subsection{Vision-based Methods} 
The performance of the localization system can be improved by the fusion of multimodal information such as inertial data, visual detections\cite{xun2023crepes,xu2020decentralized,xu2022omni,xu2024d}, and odometry\cite{cao2021relative}, among which the UWB-vision-based approach is representative. As the identity of the visual tracking target is unknown, a data association problem arises essentially. Such schemes can be divided into hardware ID-encoded-based methods and software matching-based methods depending on the ways to extract the visual identity.

ID encoding methods necessitate the arrangement of specially designed hardware devices, such as infrared LEDs\cite{xun2023crepes,yan2019active} and AprilTags\cite{olson2011apriltag}.  Yan\cite{yan2019active} equipped each drone with a distinct active infrared coded target and a monocular camera for detection, solving the relative transformation by PnP algorithm and Kalman filter. CREPES\cite{xun2023crepes} employed programmed IR LED boards and IR cameras for encoding and decoding the identity respectively, maintaining effective relative localization in dark or partially occluded scenarios.
By employing customized platforms, such methods diminished incorrect detections, but it is still encumbered by the short detection range and complex time synchronization of hardware.

For better scalability, some researchers have chosen visual detection and tracking algorithms, along with matching algorithms for data association.
Xu\cite{xu2020decentralized} utilizes convolutional neural network (CNN) detectors and MOSSE trackers to generate bearing information and retrieve the range from the depth camera, yielding visual estimated positions. The matching process involves the comparison between the positions of the final estimates and the visual estimates, as well as a predefined threshold for outlier rejection. Based on this work, their subsequent studies Omni-swarm\cite{xu2022omni} applied the Hungarian algorithm to solve the multi-robot matching issue and introduced a sparse map for global consistency. However, the above approaches overlooked the uncertainty estimation problem with the potential of erroneous matches. Additionally, depth cameras are costly and vulnerable to changes in lighting conditions, whereas our avenue employs a bearing-only matching strategy.

\section{Methodology}

Our system overview is shown in Fig. \ref{fig:pipeline}, which mainly consists of a graph match net (front-end) and a differentiable PGO module (back-end). The GNN-based front-end combines information from priors, camera detections and UWB ranges, and the resulting 3-DoF position estimates and covariances are incorporated into our differentiable back-end, finally generating the 6-DoF state estimations.


In this section, we first clarify our data association process based on GNN. Second, we describe three different constraints in PGO. Third, we introduce various loss functions for end-to-end training. Last, we illustrate the decentralized system for real-world implementation.


\begin{figure}[tp]
    \setlength{\belowcaptionskip}{-0.4cm}
    \centering
    \includegraphics[scale=0.33]{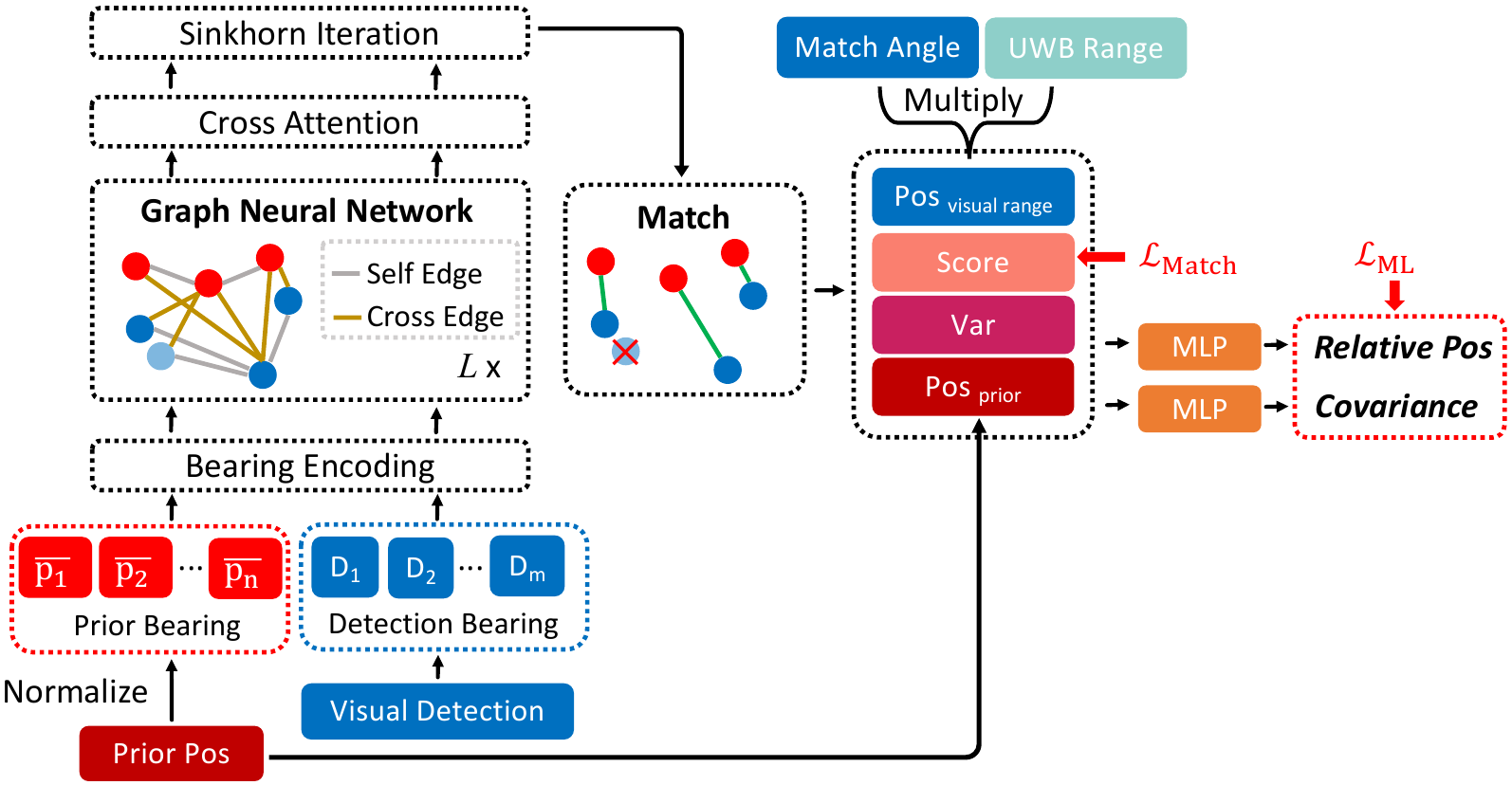}
    \caption{The graph match net front-end architecture.}
    \label{fig:front-end}
\end{figure}

\subsection{Graph Match Front-end} 
UWB rangings and priors of UAV are ID-aware, while visual detections are non-identified. Motivated by finding keypoints correspondences in the area of image feature matching, we solve the data association problem between priors and detections in a bearing-only manner.
Our network comprises the attentional graph neural network for feature aggregation, the Sinkhorn iteration for partial assignment, and the multi-layer perceptron (MLP) for estimation of positions and uncertainties. The architecture of our graph match net is illustrated in Fig. \ref{fig:front-end}.

\textbf{Attentional graph neural network:} A multi-layer graph attention neural network is utilized to encode a set of UAV bearing priors and a set of detection bearing outcomes, which are connected by self-edges and cross-edges, contributing to a deeper insight into the robot formation (intra-set) and the similarities among detection candidates (inter-set).
Each bearing of priors and detections represents a graph node. We use a shared MLP layer to project the 3-DoF bearings $\textbf{b} _{i}$ into high-dimensional space, forming the initial node embeddings $^{(0)}\textbf{f} _{i}$ that guide the network to consider spatial information.
\begin{equation}
\begin{aligned}
^{(0)}\textbf{f} _{i} = \mathit{f}_{encode}(\textbf{b} _{i})
\end{aligned}
\end{equation}

The node embeddings are then aggregated through self-attention and cross-attention, realizing comprehensive message exchange between bearings of priors and detections. Residual connections are used both within the layer and between adjacent layers. 

\begin{equation}
\setlength{\abovedisplayskip}{0.2pt}
\begin{aligned}
^{(l)}\textbf{f}^{self}_{i} = ^{(l)}\textbf{f} _{i} + ^{(l)}\mathit{f}_{self}([^{(l)}\textbf{f} _{i} \, || \, ^{(l)}\textbf{m}_{\varepsilon_{self}}]) \\
^{(l+1)}\textbf{f} _{i} = ^{(l)}\textbf{f}^{self}_{i} + ^{(l)}\mathit{f}_{cross}([^{(l)}\textbf{f}^{self}_{i} \, || \, ^{(l)}\textbf{m}_{\varepsilon_{cross}}])
\end{aligned}
\end{equation}
where $||$ denotes concatenation. $^{(l)}\textbf{f} _{i}$ is the bearing embedding of layer $l$. The message aggregation along self edges $\varepsilon_{self}$ and cross edges $\varepsilon_{cross}$ are represented by $^{(l)}\textbf{m}_{\varepsilon_{self}}$ and $^{(l)}\textbf{m}_{\varepsilon_{cross}}$ respectively. $^{(l)}\mathit{f}_{self}$ and $^{(l)}\mathit{f}_{cross}$ are MLPs, where the weights differ across layers.

After message interaction through $L$ GNN layers, the bearing embeddings of priors $\textbf{f}^{\mathcal{P}}_{i}$ and detections $\textbf{f}^{\mathcal{D}}_{j}$ are distinguishable and enriched with global information. $\mathcal{P}$ and $\mathcal{D}$ denote the set of UAV priors and detections.

\textbf{Partial assignment:} The score matrix $\mathbf S\in \mathbb{R}^{N\times M}$ can be obtained by the similarities of GNN-aggregated bearing embeddings. N and M refer to the number of drones and the maximum number of camera observations, respectively. Considering potential fake detection (the light blue drone detection in Fig. \ref{fig:pipeline}), M is larger than N. The pairwise score is computed by the dot-product:
\begin{equation}
\begin{aligned}
\textbf{S} _{i,j} = <\textbf{f}^{\mathcal{P}}_{i}, \textbf{f}^{\mathcal{D}}_{j}>
\end{aligned}
\end{equation}

In order for the network to learn to exclude mis-matching cases caused by occlusion, out-of-view and false detections, the score matrix $\mathbf S$ is further augmented to $\bar{\mathbf S}\in \mathbb{R}^{(N+1)\times (M+1)}$ by adding a dustbin row and a dustbin column for unmatched bearings. A trainable parameter is applied to represent the score of the bin row and column.

The optimal assignment can be solved by the Sinkhorn algorithm, which is differentiable, enabling end-to-end training. After several iterations, the augmented score matrix $\bar{\mathbf S}$ is reallocated subject to the constraint that the sums of rows and columns are equal to specific constant values.

The candidate match is derived according to the maximum score in each row and column. A match is considered valid only when the score exceeds a predefined threshold and both bearings of UAV priors and visual detections mutually consent to the match.

\textbf{Position prediction:} For each successfully matched UAV prior $i\in \mathcal{P}$, we construct a concatenated feature $\textbf{feat}_i$ for estimation of positions and covariances.

\begin{equation}
\begin{aligned}
\textbf{feat}_i = [^{vr}\textbf{Pos} _{i} \,|| \, ^{p}\textbf{Pos}_i \, || \, \textbf{S}_i \, || \, \textbf{Var}_i]
\end{aligned}
\end{equation}
where $^{vr}\textbf{Pos} _{i}$ is the raw visual ranging position, calculated by simply scaling the detection bearing $\textbf{b}_i \in \mathbb{R}^{3}$ with the UWB ranging $d_i \in \mathbb{R}$. $^{p}\textbf{Pos}_i$ denotes prior position. $\textbf{S}_i$ comes from the optimal matching score of row $i$ from the score matrix. $\textbf{Var}_i$ is determined by the matching probability of the detection orientation.

With the input of raw estimate, pos prior, matching score and variance, two MLPs are employed for positions and covariances prediction. Positions of unmatched drones remain as the prior, with a large covariance assigned.

\begin{equation}
\begin{aligned}
\mathbf{\hat{t}}_{i} = \mathit{f}_{pos}(\textbf{feat}_i), \: \mathbf{\hat{\Sigma}}_{i} = \mathit{f}_{cov}(\textbf{feat}_i)
\end{aligned}
\end{equation}

\subsection{Differentiable PGO Back-end}
To obtain high-precision 6-DoF poses, including the unobserved drone states, we tightly fuse sensor inputs and mutual state estimations between drones in the differentiable PGO module. Define k as the reference robot, with the relative poses of other robots to be optimized in the k coordinate system defined as:
\begin{equation}
\begin{aligned}
\boldsymbol{\chi} = [\mathbf{\hat{P}}_{1}^{k}, \mathbf{\hat{P}}_{2}^{k},..., \mathbf{\hat{P}}_{N}^{k}]
\end{aligned}
\end{equation}
where $\mathbf{\hat{P}}_{i}^{k}$ is the same as the transformation matrix 
$\begin{bmatrix}
\mathbf{\hat{R}}_{i}^{k} & \mathbf{\hat{t}}_{i}^{k}\\
 0 & \mathbf{1}
\end{bmatrix}$, $\mathbf{\hat{R}}_{i}^{k}$ refer to the rotation matrix and $\mathbf{\hat{t}}_{i}^{k}$ refer to the translational vector. To find the optimal estimation $\boldsymbol{\chi}^*$, we attempt to minimize the combined residuals:
\begin{equation}
\begin{aligned}
\boldsymbol{\chi}^* = argmin(\sum (C_{M}, C_{P}, C_{R})) 
\end{aligned}
\end{equation}
where $C_{M}, C_{P}, C_{R}$ are constraints of mutual state estimations, pose priors and range measurements respectively.

\textbf{Mutual state constraint:} The mutual observation between drone $i$ and $j$ forms a constraint edge. Since the front-end network only predicts the position, we omit the constraints on the rotation part and derive the error function in the following form:
\begin{equation}
\begin{aligned}
&\mathbf{e}_{M} = \mathbf{t}_{j}^{i} - (\mathbf{\hat{R}}_{i}^{k})^{T}(\mathbf{\hat{t}}_{j}^{k} - \mathbf{\hat{t}}_{i}^{k})\\
& C_{M} = \mathbf{e}_{M}^{T} \mathbf{\hat{\Sigma}}^{-1}_{M} \mathbf{e}_{M}
\end{aligned}
\end{equation}
where $ \mathbf{\hat{\Sigma}}^{-1}_{M}$ represents the information matrix, which is the inverse of the covariance matrix,   where the diagonal elements are formed by the uncertainties predicted by the preceding network.

\textbf{Pose prior constraint:} When the number of mutual observations decreases, the optimization problem may become ill-conditioned, due to which we incorporate prior pose constraints to avoid degradation.
\begin{equation}
\begin{aligned}
&\mathbf{e}_{P} = \mathbf{P}_{i}^{k} - \mathbf{\hat{P}}_{i}^{k} \\
& C_{P} = \mathbf{e}_{P}^{T}  \mathbf{\hat{\Sigma}}^{-1}_{P} \mathbf{e}_{P}
\end{aligned}
\end{equation}
The diagonal entries of the covariance matrix $ \mathbf{\hat{\Sigma}}_{P}$ are likewise predicted by the network.

\textbf{UWB ranging constraint:} The UWB measurements establish pairwise connections between drones in the cluster.
\begin{equation}
\begin{aligned}
&\mathbf{e}_{R} = \textit{d}_{ij} - \parallel \mathbf{\hat{t}}_{i}^{k} - \mathbf{\hat{t}}_{j}^{k} \parallel_ 2\\
& C_{R} = \mathbf{e}_{R}^{T}  \mathbf{\hat{\Sigma}}^{-1}_{R} \mathbf{e}_{R}
\end{aligned}
\end{equation}
where $\textit{d}_{ij}$ is the distance measurement between drone $i$ and $j$. The covariance of UWB ranging $ \mathbf{\hat{\Sigma}}_{R}$ is predefined.

We use second-order Levenberg-Marquardt (LM) algorithm and Cholmod sparse solver in Theseus\cite{pineda2022theseus} to solve the nonlinear optimization problem. The gradient of each iteration is stored and backpropagated to facilitate the training of the front-end network.

\subsection{Loss Functions}
To guide the network to produce robust matches, reliable covariance distributions and accurate state estimations, three different loss functions are applied in our training, balanced by factors $\lambda_1, \lambda_2$:
\begin{equation}
\begin{aligned}
\mathcal{L} = \mathcal{L}_{Match} + \lambda_1 \mathcal{L}_{ML} + \lambda_2 \mathcal{L}_{Pose}
\end{aligned}
\end{equation}

\textbf{Match loss:} For the matching item, we supervise the augmented score matrix $\bar{\mathbf S}$ after Sinkhorn iterations:
\begin{equation}
\begin{aligned}
\mathcal{L}_{Match} = -\sum_{(i,j)\in \pi} \bar{\mathbf S}_{i,j} - \sum_{i\in \mu} \bar{\mathbf S}_{i,M}
\end{aligned}
\end{equation}
where $\pi$ is the set of matching bearings, $\mu$ denotes the set of unmatched UAV priors, both of which come from the ground truth labels. The column $M$ in $\bar{\mathbf S}_{i,M}$ represents the dustbin column, storing the cases where camera observations are lost. The first item encourages the net to amplify the scores of correct matches, while the second term drives the network to exclude incorrect ones.

\textbf{Maximum likelihood loss:} According to the form of a multivariate Gaussian distribution, we define a negative log-likelihood covariance loss as follows:
\begin{equation}
\begin{split}
\mathcal{L}_{ML} = &\frac{1}{(N+1)*N} \sum_{(i,j)\in \mathcal{S}, j\ne i} (\lambda_{det}log(det(\mathbf{\hat{\Sigma}}^i_j))\\
                    &+ (\mathbf{t}_{j}^{i} - \mathbf{\hat{t}}_{j}^{i})^T \, \mathbf{\hat{\Sigma}}^{i^{-1}}_{j} \, (\mathbf{t}_{j}^{i} - \mathbf{\hat{t}}_{j}^{i}))
\end{split}
\end{equation}
where $\mathcal{S}$ is the set of UAV swarm, $\mathbf{\hat{\Sigma}}^i_j \in \mathbb{R}^{3\times 3}$ is the predicted covariance matrix, $\mathbf{\hat{t}}_{j}^{i}$ is the 3-DoF position estimate of drone $j$ with respect to drone $i$, while $\mathbf{t}_{j}^{i}$ is the ground truth.


\textbf{Pose loss:} Apart from the aforementioned two losses directly applied to the front-end net outputs, we also define the Mean Square Error (MSE) pose loss for final relative localization after graph optimization:
\begin{equation}
\begin{split}
\mathcal{L}_{Pose} = &\frac{1}{(N+1)*N} \sum_{(i,j)\in \mathcal{S}, j\ne i} (\parallel \mathbf{t}_{j}^{i} - \mathbf{\hat{t}}_{j}^{i} \parallel^2 \\
                     & + \lambda_{q} \parallel \mathbf{q}_{j}^{i} - \mathbf{\hat{q}}_{j}^{i} \parallel^2 )
\end{split}
\end{equation}
where $\mathbf{\hat{t}}_{j}^{i}$ and $\mathbf{\hat{q}}_{j}^{i}$ denote the estimation of translation and quaternion part, $\mathbf{t}_{j}^{i}$ and $\mathbf{q}_{j}^{i}$ are their corresponding ground truths.


\begin{figure}[tp]
    \setlength{\belowcaptionskip}{-0.4cm}
    \centering
    \includegraphics[scale=0.35]{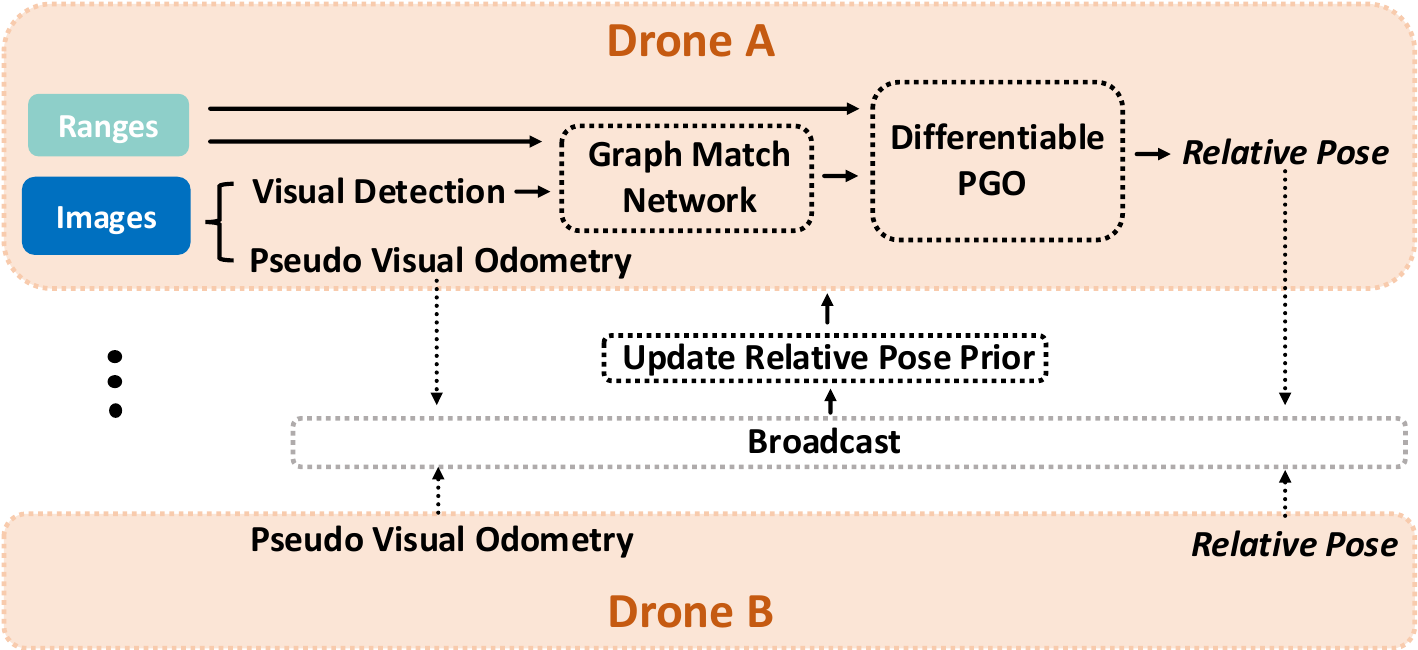}
    \caption{Decentralized system diagram.}
    \label{fig:decentralized}
\end{figure}

\subsection{Decentralized System}
For the purpose of high-performance deployment in real environments, we have realized a decentralized system (Fig. \ref{fig:decentralized}). The system is built upon the ROS communication framework, the graph match front-end network is implemented on LibTorch, and the optimization back-end is solved by Ceres Solver.

In the simulation experiments, the visual detection direction is computed based on the orientation and positional relationship between two drones within the field of view. In the physical experiments, we use the infrared hardware module of CREPES for recognition (without ID extraction).
As drones usually fly at high speeds, we introduce pseudo visual odometry (PVO) to aid in updating the relative pose prior between consecutive frames, which is derived from the ground truth with considerable noise. The relative pose priors are updated by the PVO and the mutual state estimation among other neighboring robots. 
Only the PVO and the optimized relative poses in the local frames are shared among the drones, leading to a low communication load.


\section{Experiment}
We carry out experiments on both simulation datasets and self-collected real-world datasets to verify the accuracy and robustness of our proposed method, covering both line-of-sight (LOS) and non-line-of-sight (NLOS) scenarios. The details of our experiment datasets are listed in TABLE. \ref{table:datasets}.

\begin{table}[bp]
\setlength{\belowcaptionskip}{-0.3cm}
\caption{Experiment Datasets.}
\label{table:datasets}
\renewcommand\arraystretch{1.2}
\centering
\begin{tabular}{ccccc}
\hline
Dataset             & Drone Num & Detection Num & Occlusion & Traj Len \\ \hline
\textit{Sim-Forest} & 4/8/12/16        & 0$\sim$8/12/16/20     & \checkmark       & 52.23m   \\
\textit{Real-LOS}   & 5         & 0$\sim$8      & \ding{55}        & 20.64m   \\
\textit{Real-NLOS}  & 5         & 0$\sim$8      & \checkmark       & 20.63m   \\ \hline
\end{tabular}
\end{table}

\subsection{Experimental Settings and Datasets}
A laptop equipped with a 13th Gen Intel Core i9-13900HX CPU and Nvidia RTX 4060 GPU is used to train and validate our neural network. We employ a 4-layer GNN network to aggregate the bearing features. The number of Sinkhorn iterations is set to 100. We use the Adam optimizer for training, with a learning rate of 1e-4 and a weight decay coefficient of 5e-4. The front-end network undergoes pretraining to generate accurate estimates of positions and uncertainties, which takes less than 50 epochs.

At train time, we add noise with a standard deviation of 0.1m along each axis to the ground truth position from 0.1 seconds earlier, indicating the pos prior. At test time, the prior comes from the previous prediction plus the relative odometry increment. In situations where all camera detections or UWB observations are unavailable, these instances will be omitted during training, while for testing, the position prior will be simply updated by odometry and fed into the PGO, bypassing the graph match network.

\begin{figure*}[htbp]
    \centering
    \includegraphics[scale=0.45]{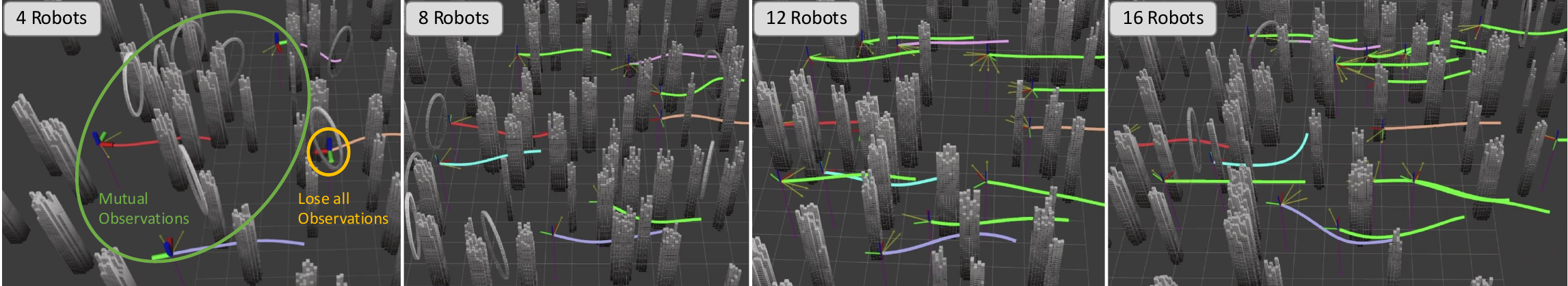}
    \caption{The simulated random forest environment with varying robot counts. In the four-robot scenario, the three robots with mutual observations are circled by green ellipses, while the robot that lost all observations due to occlusion is highlighted by a yellow ellipse. As the number of robots increases, occlusions are more likely to occur.}
    \label{fig:sim}
\end{figure*}

\textbf{Simulation scenes:} We conduct simulation experiments in a random forest environment filled with diverse obstacles like trees. A group of drones in a circular formation traverses the $70m\times30m\times3m$ forest. The simulated scenes with varying numbers of robots are presented in Fig. \ref{fig:sim}.
Each UAV has a 180-degree field of view (FOV), and the visibility is not only affected by occlusion but also influenced by both the camera orientation of the observing drones and the direction of the IR LEDs on the drones being observed. To validate the robustness of our proposed matching method, we also randomly generate erroneous visual detections with a probability exceeding 40\%.

\textbf{Real-world scenes:} In the physical experiments, we adopted the same hardware as CEREPS for data acquisition, while excluding its ID extraction and IMU modules. The camera features a fisheye lens with a 185-degree FOV, and the UWB module is from the NoopLoop DW1000 series. As the experiments are carried out indoors, we employ MCS to obtain the ground truth. The effectiveness of our proposal is demonstrated in both occlusion and non-occlusion scenarios.

\subsection{Baselines}
We choose two baselines for comparison with our approach.

\textbf{PVO:} The odometry of each robot is derived by adding noise perturbations to the ground truth. The noise consists of a translation disturbance with a standard deviation of 0.1m and a rotation disturbance with a standard deviation of 1.0 degree, applied every 0.1s of odometry. Note that the noise magnitude is consistent with that of the other PVO-aided methods. The relative pose estimates calculated by the odometry are further passed into the PGO for optimization, with no inter-robot observations involved.

\textbf{Simple Match:} The data association is performed according to the closest direction between the bearings of UAV priors and visual detections. A tunable threshold is used to discard matches with large directional differences. For successfully matched drones, their relative positions are obtained by multiplying the camera directions with the UWB rangings. The covariances have a negative correlation with the cosine similarities of the matching bearings, contributing to more robust optimization compared to hard matching. This method also leverages PVO to update the state priors and enhances the overall accuracy through graph optimization.

\subsection{Multi-Robot Localization Accuracy}

\begin{table}[bp]
\setlength{\belowcaptionskip}{-0.3cm}
\caption{RPE evaluation under simulated and real-world datasets. Numbers in parentheses following the scene name denote the quantities of robots. Values in bold are the best.}
\label{table:rpe}
\renewcommand\arraystretch{1.2}
\centering
\begin{tabular}{cclclcl}
\hline
\multirow{2}{*}{Method}       & \multicolumn{6}{c}{Position RMSE (m)}                                                            \\ \cline{2-7} 
             & \multicolumn{2}{c}{\textit{Sim-Forset}(16)} & \multicolumn{2}{c}{\textit{Real-LOS}(5)}   & \multicolumn{2}{c}{\textit{Real-NLOS}(5)}  \\ \hline
PVO & \multicolumn{2}{c}{6.067}          & \multicolumn{2}{c}{2.243}          & \multicolumn{2}{c}{1.445}          \\ 
Simple Match          & \multicolumn{2}{c}{0.198}          & \multicolumn{2}{c}{0.108}          & \multicolumn{2}{c}{0.498}          \\ 
Ours         & \multicolumn{2}{c}{\textbf{0.144}} & \multicolumn{2}{c}{\textbf{0.090}} & \multicolumn{2}{c}{\textbf{0.129}} \\ \hline
\end{tabular}
\end{table}

\begin{figure*}[htbp]
\setlength{\belowcaptionskip}{-0.2cm}
    \centering
    \includegraphics[scale=0.55]{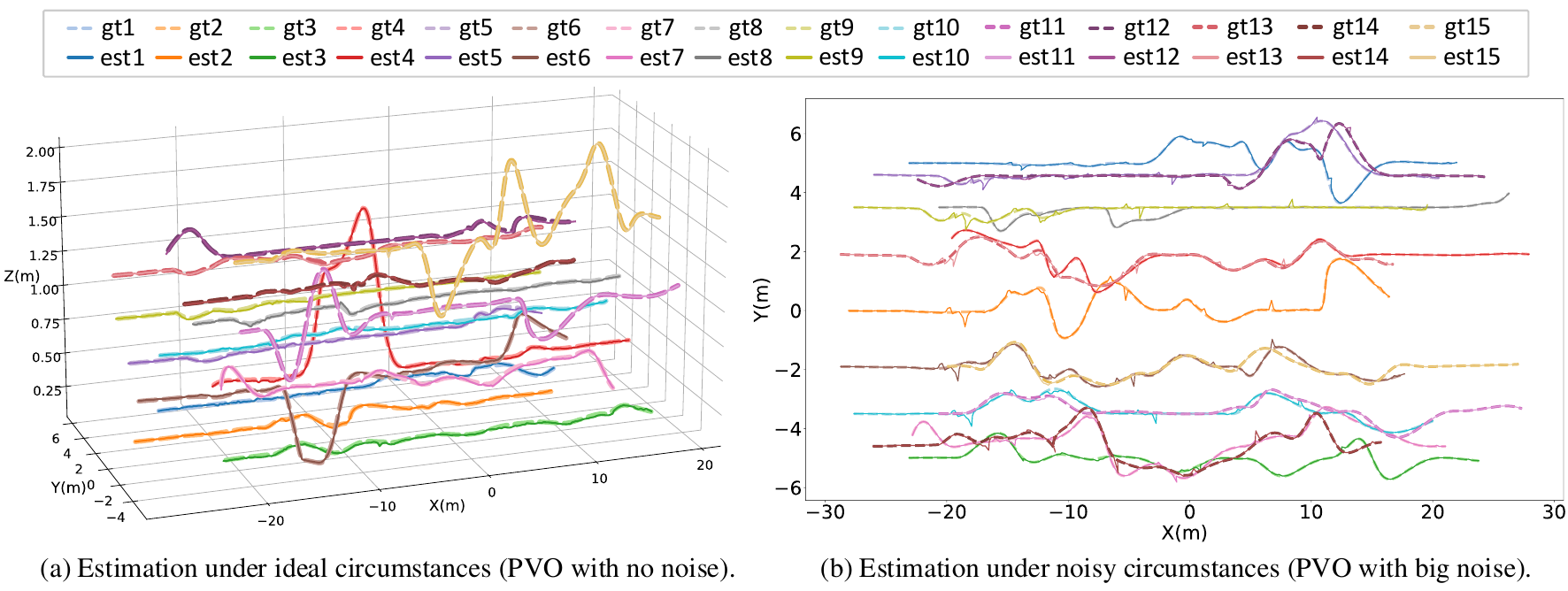}
    \caption{The estimated trajectories of other 15 drones on simulated forest environment.}
    \label{fig:traj-sim}
\end{figure*}

\begin{figure}[htbp]
\setlength{\belowcaptionskip}{-0.5cm}
    \centering
    \includegraphics[width=.49\textwidth]{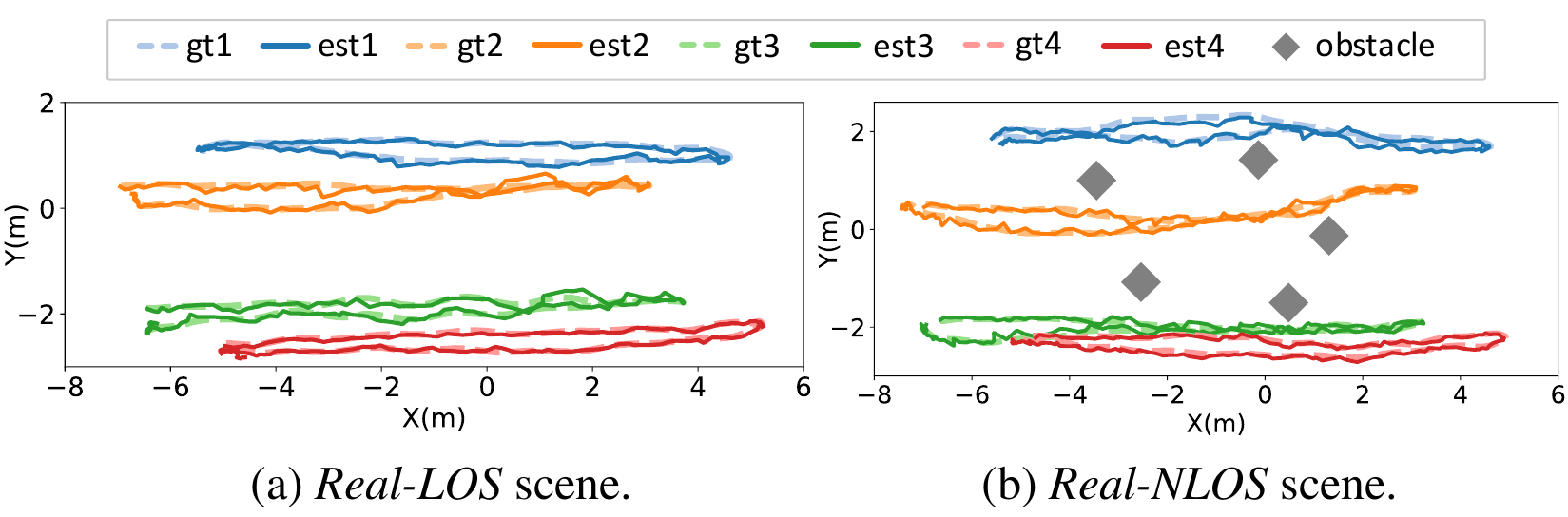}
    \caption{The estimated trajectories of other 4 drones under noisy circumstances (PVO with big noise) in real-world environment.}
    \label{fig:traj-real}
\end{figure}

\begin{figure}[tp]
\setlength{\abovecaptionskip}{-0.1cm}
\setlength{\belowcaptionskip}{-0.5cm}
\centering
\subfigure[RPE of ours.]{
\includegraphics[width=.22\textwidth]{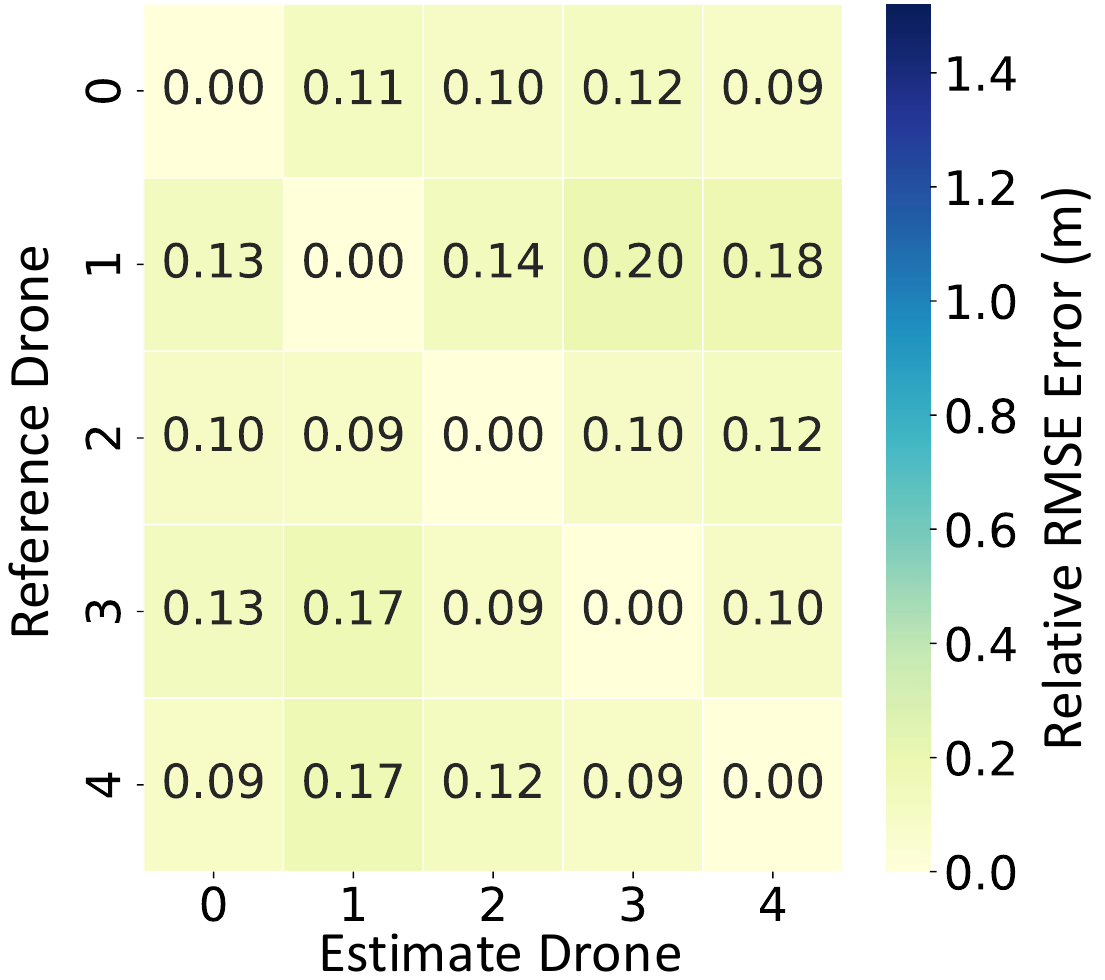}
}
\subfigure[RPE of Simple Match.]{
\includegraphics[width=.22\textwidth]{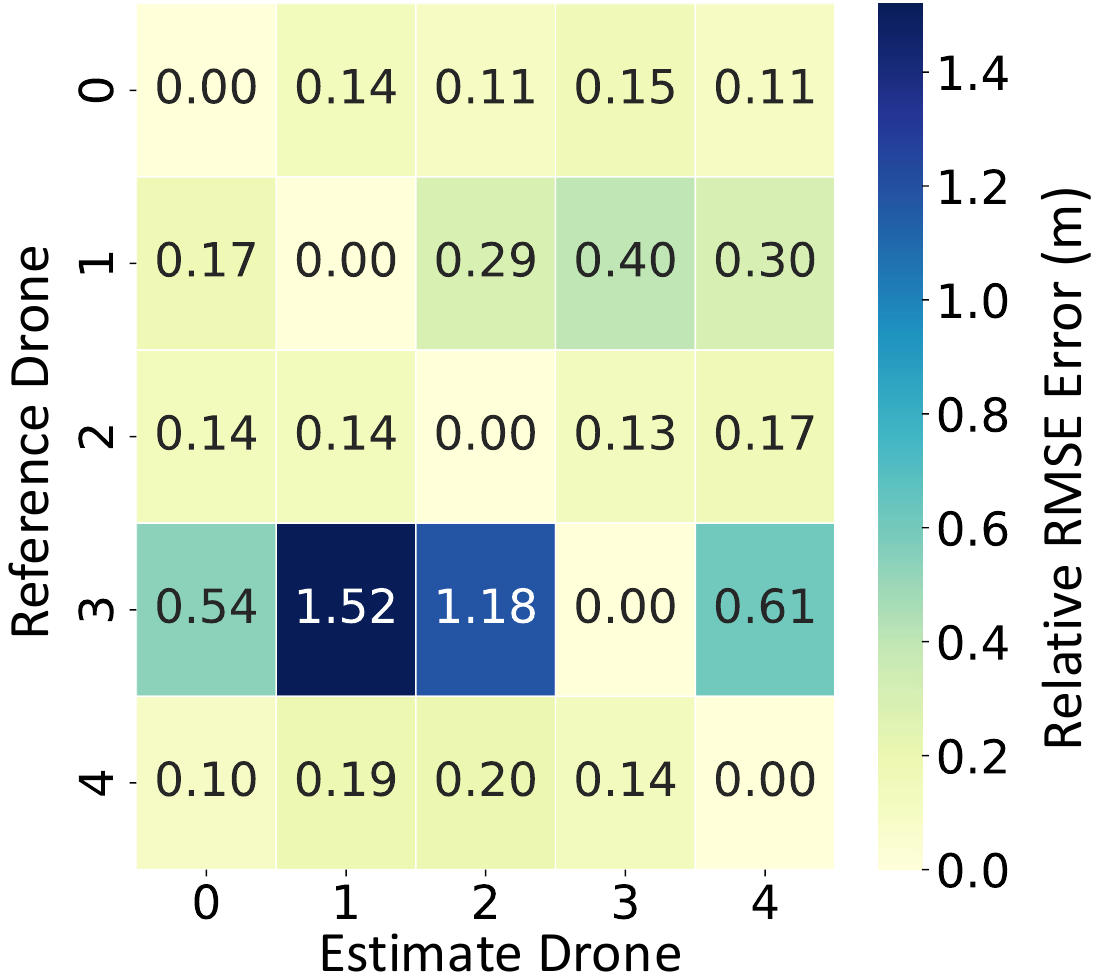}
}
\caption{The RPE heat map w.r.t each drone on \textit{Real-NLOS}.}
\label{fig:heat_map}
\end{figure}

For ease of visualization, estimated trajectories of other drones are transformed by adding the relative poses on the ground truth of drone 0. 
The estimations under ideal conditions are shown in Fig. \ref{fig:traj-sim}(a), achieving the RMSE error of 3.9cm. 
In actual conditions, odometry suffers from accumulated drift, and the estimations with considerable noise added to the PVO are presented in Fig. \ref{fig:traj-sim}(b) and Fig. \ref{fig:traj-real}. We evaluate 3D Relative Positioning Error (RPE) by RMSE in TABLE. \ref{table:rpe} and the RPE of dataset \textit{Real-NLOS} with respect to each robot estimated by our approach and Simple Match are depicted by heat map in Fig. \ref{fig:heat_map}.
Notably, all error metrics are calculated in the local frame of each robot, while trajectories of in the world coordinate frame are used only for visualization.

\textbf{Resistance to noise interference:} The noise we introduce into the odometry is substantially greater than the errors of most state-of-the-art visual odometry methods. As shown in TABLE. \ref{table:rpe}, PVO quickly diverges in all three scenarios, leading to large errors under the influence of noise. With the same noise level, the predicted trajectories of our approach exhibit some fluctuation when inter-observations decrease, but it quickly stabilizes once visibility is regained, maintaining valid and robust localization throughout the whole flight.

\textbf{Occlusion resistance:} As indicated in TABLE. \ref{table:rpe}, the Simple Match based on the closest bearing matching and the fixed threshold filter achieves comparable performance to our proposed method in non-occlusion environments (\textit{Real-LOS}). However, a predefined threshold necessitates a trade-off between precision and recall rate. A lenient threshold parameter may result in more false matches, while a strict threshold reduces the matching recall rate. In occlusion scenarios (\textit{Sim-Forest} and \textit{Real-NLOS}), the Simple Match has fewer valid matches,  relying solely on the noisy odometry to continuously update the state priors, causing significant performance degradation. In contrast, our graph match network focuses on the overall similarity between the set of UAV priors and visual detections in a global view, producing more successful matches and exhibiting stronger resistance to occlusion.

\begin{table*}[!ht]
\setlength{\abovecaptionskip}{-0.02cm}
\caption{Matching results on simulated and real-world datasets. Numbers in parentheses following the scene name denote the quantities of robots. F1 scores in bold are the best.}
\label{table:match}
\renewcommand\arraystretch{1.2}
\centering
\begin{tabular}{ccccccccccccc}
\hline
\multirow{3}{*}{Method} & \multicolumn{12}{c}{Datasets}                                                                                                                                                                                        \\ \cline{2-13} 
                        & \multicolumn{3}{c|}{\textit{Sim-Forset} (8)}                & \multicolumn{3}{c|}{\textit{Sim-Forset} (16)}                & \multicolumn{3}{c|}{\textit{Real-LOS} (5)}                  & \multicolumn{3}{c}{\textit{Real-NLOS} (5)} \\ \cline{2-13} 
                        & P       & R       & \multicolumn{1}{c|}{F1}             & P       & R       & \multicolumn{1}{c|}{F1}             & P       & R       & \multicolumn{1}{c|}{F1}             & P         & R        & F1              \\ \hline
Simple Match@0.9      & 76.49\% & 99.45\% & \multicolumn{1}{c|}{0.865}          & 58.96\% & 97.54\% & \multicolumn{1}{c|}{0.735}          & 88.74\% & 98.85\% & \multicolumn{1}{c|}{0.935}          & 88.59\%   & 99.78\%  & 0.939           \\ 
Simple Match@0.99     & 95.11\% & 98.50\% & \multicolumn{1}{c|}{0.968}          & 87.28\% & 95.94\% & \multicolumn{1}{c|}{0.914}          & 97.32\% & 92.41\% & \multicolumn{1}{c|}{0.948}          & 99.11\%   & 96.14\%  & 0.976           \\ 
Ours                    & 97.87\% & 98.58\% & \multicolumn{1}{c|}{\textbf{0.982}} & 95.92\% & 91.14\% & \multicolumn{1}{c|}{\textbf{0.935}} & 96.15\% & 98.33\% & \multicolumn{1}{c|}{\textbf{0.972}} & 97.47\%   & 99.51\%  & \textbf{0.985}  \\ \hline
\end{tabular}
\end{table*}


\subsection{Ablation Study}
In this section, we will examine the strengths of front-end learning graph networks compared to Simple Match, as well as the enhancement of accuracy achieved through back-end graph optimization.

\begin{figure}[tp]
\setlength{\belowcaptionskip}{-0.6cm}
    \centering
    \includegraphics[scale=0.35]{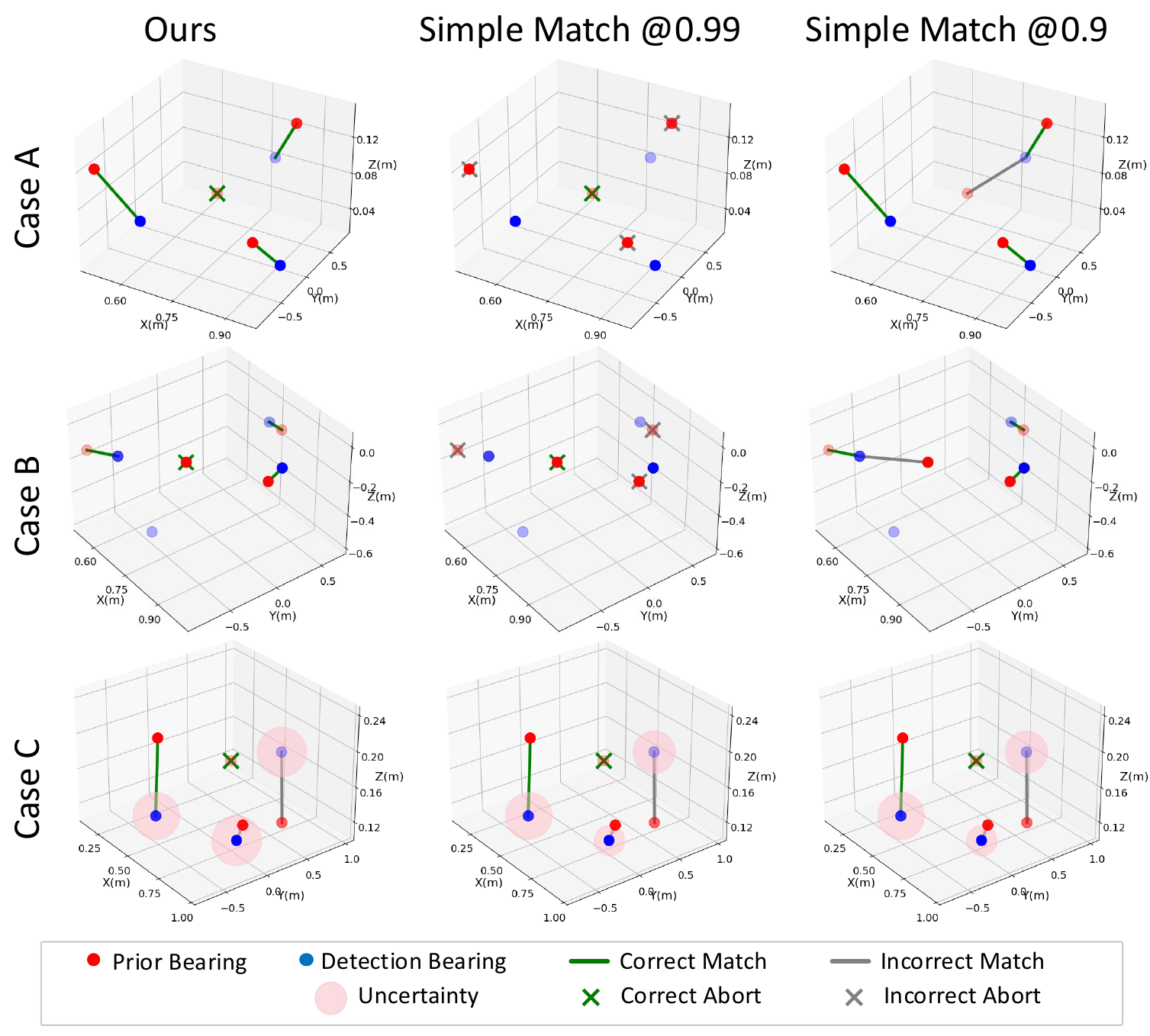}
    \caption{Data association cases. We perform a comparison between our graph matching network and Simple Match with distinct filtering thresholds. In challenging Case C, the uncertainty of the matches is represented by pink spheres, where a larger radius indicates a higher uncertainty.}
    \label{fig:match_case}
\end{figure}

\textbf{Graph match front-end:} We evaluate the graph matching network in comparison with the Simple Match under different threshold settings, measuring the matching precisions, recall rates, and F1 scores across multiple datasets. All metrics are presented in Table. \ref{table:match}. Simple Match@0.9 indicates that only matches with a cosine similarity greater than 0.9 between the UAV priors and the detection bearings are considered valid. Likewise, Simple Match@0.99 stands for a stricter threshold parameter. The F1 score takes both the matching precision and recall rate into account, and our proposed method achieves the highest F1 score across all datasets, which highlights that our model strikes a good balance between incorporating anonymous observations and excluding incorrect matches. A few representative cases are displayed in Fig. \ref{fig:match_case}. In Case A, there are four drones to be matched and three camera observations. Our method successfully matches all observed drones and excludes the one that is not observed, while Simple Match@0.99 rejects all matches and Simple Match@0.9 improperly associates two different drones with the same detection bearing based on the closest direction matching. A similar situation also occurs in Case B, where there are three valid camera observations and one erroneous camera detection. In the more complex and challenging Case C, all three methods encounter two incorrect matches, while our approach applies higher uncertainty to these mismatches, reducing their detrimental effect on the subsequent optimization process.

\textbf{PGO back-end:} To quantify the precision improvement with the incorporation of PGO, we conduct ablation experiments in a centralized inference strategy, ensuring identical communication conditions. The PVO is perturbed by noise and camera observations are interfered with random fake detections. The results listed in Table. \ref{table:pgo} demonstrate that the incorporation of PGO efficiently mitigates the localization error of either Simple Match or our learned front-end by over 32\% across all scenarios.

\begin{table}[tp]
\caption{Ablation results of PGO. The errors are evaluated by RPE RMSE.}
\label{table:pgo}
\renewcommand\arraystretch{1.2}
\setlength{\tabcolsep}{3pt}
\centering
\begin{tabular}{ccc|cccccc}
\hline
\multicolumn{3}{c|}{Module}                                                       & \multicolumn{6}{c}{Scene}                                                                                                                            \\ \hline
Simple                    & Learned                   & PGO                       & \multicolumn{2}{c|}{\textit{Sim-Forest}(16)}             & \multicolumn{2}{c|}{\textit{Real-LOS}(5)}               & \multicolumn{2}{c}{\textit{Real-NLOS}(5)} \\ \hline
\checkmark & \ding{55}    & \ding{55}    & 1.572 & \multicolumn{1}{c|}{\multirow{2}{*}{$\downarrow$46.8\%}} & 0.293 & \multicolumn{1}{c|}{\multirow{2}{*}{$\downarrow$42.0\%}} & 0.453     & \multirow{2}{*}{$\downarrow$55.8\%}    \\
\checkmark & \ding{55}    & \checkmark & 0.836 & \multicolumn{1}{c|}{}                        & 0.170 & \multicolumn{1}{c|}{}                        & 0.200     &                            \\ \hline
\ding{55}    & \checkmark & \ding{55}    & 1.280 & \multicolumn{1}{c|}{\multirow{2}{*}{$\downarrow$89.2\%}} & 0.217 & \multicolumn{1}{c|}{\multirow{2}{*}{$\downarrow$32.9\%}} & 0.247     & \multirow{2}{*}{$\downarrow$35.2\%}    \\
\ding{55}    & \checkmark & \checkmark & 0.138 & \multicolumn{1}{c|}{}                        & 0.126 & \multicolumn{1}{c|}{}                        & 0.160     &                            \\ \hline
\end{tabular}
\end{table}

\begin{figure}[tp]
\setlength{\belowcaptionskip}{-0.55cm}
\centering
\subfigure[Decrease of combination loss.]{
\includegraphics[width=.22\textwidth]{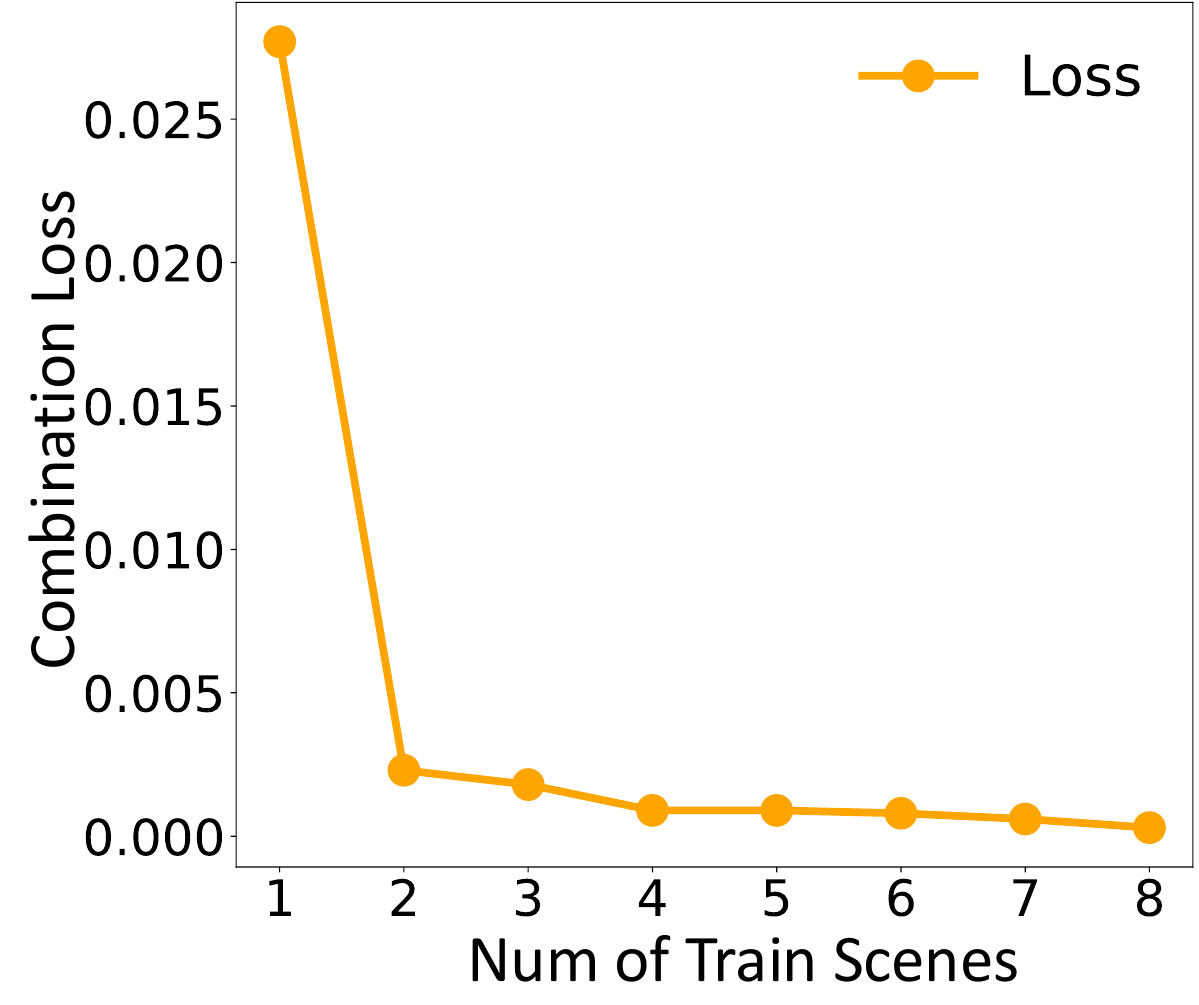}
}
\subfigure[Decrease of RPE.]{
\includegraphics[width=.22\textwidth]{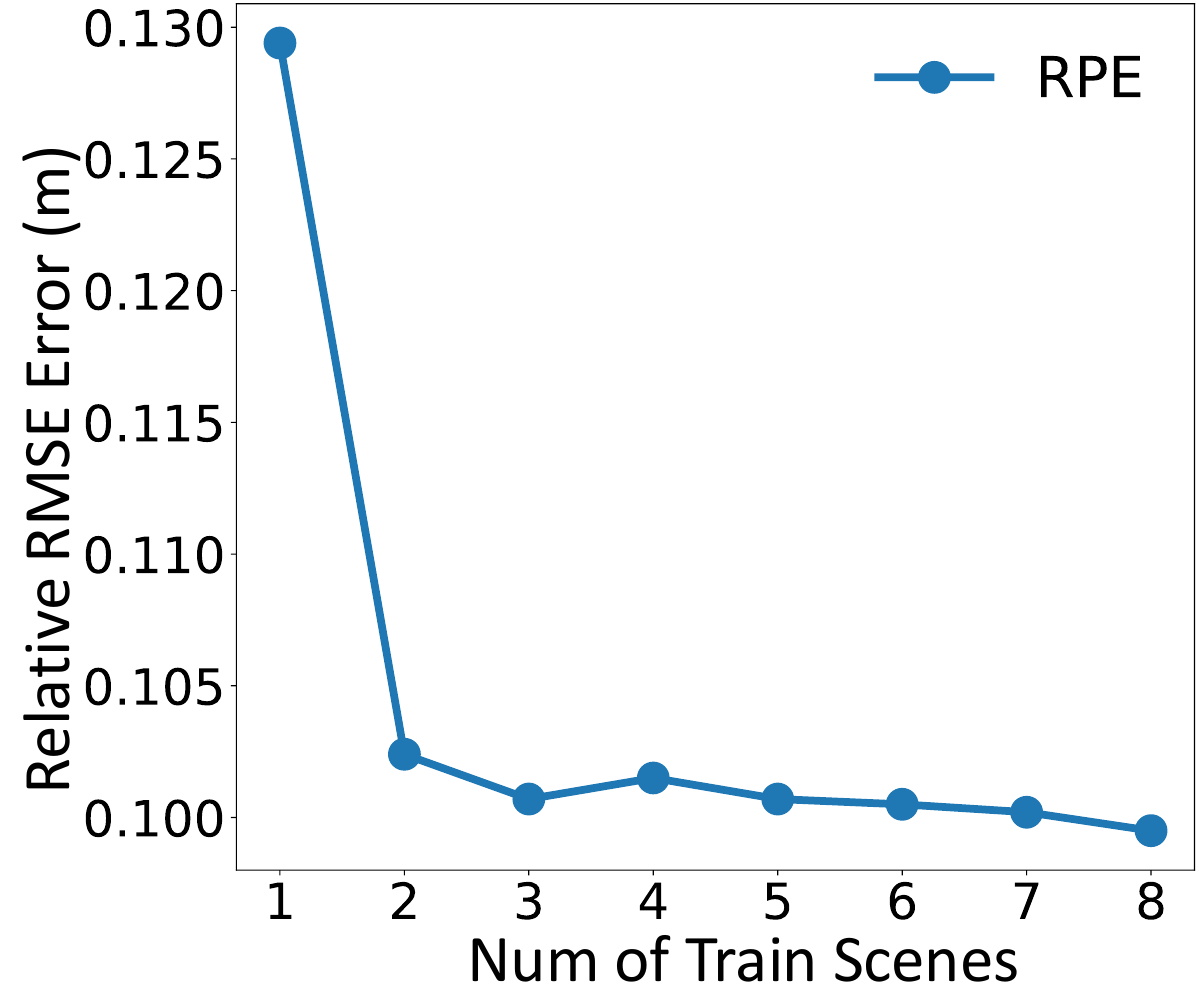}
}
\caption{The decrease of combination loss and RPE w.r.t number of training scenes.}
\label{fig:multi-scene}
\end{figure}

\subsection{Network Generalization Ability Analysis}
In this section, we will investigate the generalization of the neural network across varying training set sizes, different numbers of robots, and simulation-to-real model transfer experiments.

\textbf{Training scene numbers:} We use varying numbers of scenes for training, assessing its inference performance in real-world occlusion environments. The combined loss and RPE error are presented in Fig. \ref{fig:multi-scene}. The model reaches performance comparable to the optimal model after being trained on only two scenes.

\begin{figure}[tp]
    \setlength{\belowcaptionskip}{-0.2cm}
    \centering
    \includegraphics[width=.45\textwidth]{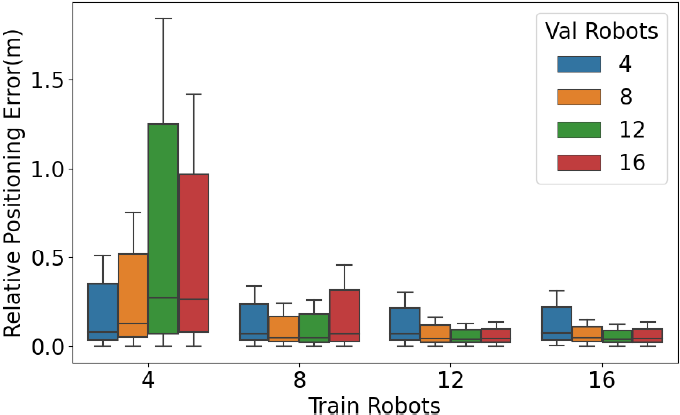}
    \caption{RPE box plot of varying robot numbers experiment. The whisker length in the box plot is defined as 0.5 times the interquartile range (IQR).}
    \label{fig:box}
\end{figure}

\textbf{Robot numbers:} We train and test in simulated scenes with different numbers of robots, validating the model's generalization performance across varying robot nodes. To eliminate the differences in training set size caused by varying numbers of robots, we scale the training size by corresponding multiples. For example, the model for 16 robots is trained with 4 scenes, while the model for 8 robots is trained with 8 scenes. As illustrated by the error box plot in Fig. \ref{fig:box}, networks trained with a larger number of robots achieve higher accuracy on all datasets. An increased number of robots enables the GNN to learn more intricate topological structures, offer more reasonable uncertainty estimates, and generalize to scenes with different robot quantities.

\textbf{Sim-to-real:} To evaluate the network’s adaptability from simulation models to real-world data, the simulation model trained on \textit{Sim-Forest} sequences is validated in real-world occlusion and non-occlusion environments. The accuracy gap between the simulation model and the real model is less than 1cm, as shown in Table. \ref{table:sim2real}.

\begin{table}[tp]
\caption{The accuracy of simulation model and real model on real-world scenarios.}
\label{table:sim2real}
\renewcommand\arraystretch{1.2}
\setlength{\tabcolsep}{5pt}
\centering
\begin{tabular}{cccl|clcl}
\hline
\multirow{2}{*}{Model} & \multicolumn{3}{c|}{Training Setting}                       & \multicolumn{4}{c}{Position RMSE (m)}                                          \\ \cline{2-8} 
                        & Scene            & \multicolumn{2}{c|}{Num of Robot / Cam} & \multicolumn{2}{c}{\textit{Real-LOS}} & \multicolumn{2}{c}{\textit{Real-NLOS}} \\ \hline
Sim               & Sim-Forest        & \multicolumn{2}{c|}{16 / 20}            & \multicolumn{2}{c}{0.098}             & \multicolumn{2}{c}{0.136}              \\ 
Real              & Real-world        & \multicolumn{2}{c|}{5 / 8}              & \multicolumn{2}{c}{0.090}             & \multicolumn{2}{c}{0.129}              \\ \hline
\end{tabular}
\end{table}

\section{Conclusions}
In this work, we propose an end-to-end learning visual-range framework in multi-agent relative localization system. The learnable front-end solves the data association problem through multiplex attentional graph neural network, facilitating reliable matching and uncertainty prediction. The differentiable PGO back-end collects mutual estimations and boost the overall precision. 

In future work, we will consider replacing PVO with a real visual odometry system and integrating multi-frame observations into the network and optimization framework, which will further enhance the system's robustness and adaptability.

\bibliographystyle{IEEEtran}
\bibliography{Mybib}

\end{document}